# Visual Tracking using Particle Swarm Optimization

J.R.Siddiqui and S.Khatibi,
Department of Computing, Blekinge Institute of Technology, Sweden.

## ABSTRACT

*The problem of robust extraction of visual odometry from a sequence of images obtained by an eye in hand camera configuration is addressed. A novel approach toward solving planar template based tracking is proposed which performs a non-linear image alignment for successful retrieval of camera transformations. In order to obtain global optimum a bio-metaheuristic is used for optimization of similarity among the planar regions. The proposed method is validated on image sequences with real as well as synthetic transformations and found to be resilient to intensity variations. A comparative analysis of the various similarity measures as well as various state-of-art methods reveal that the algorithm succeeds in tracking the planar regions robustly and has good potential to be used in real applications.*

## 1. INTRODUCTION

Accurate relative position estimation by keeping track of salient regions of a scene can be considered to be the core functionality of a navigating body such as a mobile robot. These salient regions are often referred to as "Landmarks" and the process of position estimation and registration of landmarks on a local representation space (i.e. a Map) is called SLAM (Simultaneous Localization and Mapping). The choice of landmarks and their representation depends on the environment as well as the configuration of a robot. In the case of vision based navigation, feature oriented land-marking is often employed, where features can be represented in many ways (e.g. by points, lines, ellipses and moments) [1]. Such techniques either do not exploit rigidity of the scene [2]–[4] or geometrical constraints are loosely coupled by keeping them out of the optimization process [5]–[7]. These techniques can therefore have inaccurate motion estimation due to small residual errors incurred in each iteration which make motion estimations inaccurate as these errors get accumulated. In order to mitigate this, an additional correction step is often added which either exploits a robot's motion model to predict the future state using an array of extended Kalman-Filters [8] or minimizes the integrated error calculated over a sequence of motion [9].

Generally, feature-oriented ego-motion estimation approaches [10], [11] follow three main steps; feature extraction, correspondence calculation and motion estimation. The extracted features are mostly sparse and the process of extraction is decoupled from motion estimation. Sparsification and decoupling makes a technique less computationally expensive and also allows it to handle large displacements in subsequent images, however accuracy suffers when the job is to localize a robot and map the environment for a longer period of time. Since finding correspondences is itself an error-prone task, a large portion of the error is introduced in a very early phase of motion estimation.

There is another range of methods that utilize all pixels of an image region when calculating camera displacement by aligning image regions and hence enjoy higher accuracy due to exploitation of all possible visual information present in the segments of a scene [12]. These methods are termed "direct image alignment" based approaches for motion estimation because they do not have feature extraction and correspondence calculation steps and work directly on image patches. Direct methods are often avoided due to their computational expense which overpowers the benefits of accuracy they might provide, however an intelligent selection of the important parts of the scene that are rich in visual information can provide a useful way of dealing with the issue [13]. In addition to being direct in their approach, such methods can also better exploit the geometrical structure of the environment by including rigidity constraints early in the optimization process. The use of all visual information in a region of an image and keeping track of gain or loss in subsequent snapshots of a scene is also relevant, since it is the way some biological species navigate. For example, there are evidences that desert ants use the amount of visual information which is common between a current image and a snapshot of the ant pit to determine their way to the pit [14].

An important step in a direct image alignment based motion estimation approach is the optimization of similarity among image patches. The major optimization technique that is extensively used for image alignment is gradient descent although a range of algorithms (e.g. Gauss-Newton, Newton-Raphson and Levenberg-Marquardt [9], [15]) are used for calculation of a gradient descent step. Newton's method provides a high convergence rate because it is based on second order Taylor series approximation, however, Hessian calculation is a computationally expensive task. Moreover, a Hessian can also be indefinite, resulting in convergence failure. These methods perform a linearization of the non-linear problem which can then be solved by linear-least square methods. Since these methods are based on gradient descent, and use local descent to determine the direction of an optimum in the search space, they have a tendency to get stuck in the local optimum if the objective function has multiple optima. There are, however, some bio-inspired metaheuristics that mimic the behavior of natural organisms (e.g. Genetic Algorithms (GAs) and Particle Swarm Optimization (PSO) [16]–[18] ) or the physical laws of nature to cater this problem [19]. These methods have two common functionalities: exploration and exploitation. During an exploration phase, like any organism explores its environment, the search space is visited extensively and is gradually reduced over a period of iterations. The exploitation phase comes in the later part of a search process, when the algorithm converges quickly to a local optimum and the local optimum is accepted as the global best solution. This two-fold strategy provides a solid framework for finding the global optimum and avoiding the local best solution at the same time. In this case, PSO is interesting as it mimics the navigation behavior of swarms, especially colony movement of honeybees if an individual bee is represented as a particle which has an orientation and is moving with a constant velocity. Arbitrary motion in the initial stage of the optimization process ensures better exploration of the search space and a consensus among the particles reflects better convergence.

In this paper, the aim is to solve the problem of camera motion estimation by directly tracking planar regions in images. In order to learn an accurate estimate of motion and to embed the rigidity constraint of the scene in the optimization process, a PSO based camera tracking is performed which uses a non-linear image alignment based approach for finding the displacement of camera within subsequent images. The major contributions of the paper are: a) a novel approach to planar template based camera tracking technique which employs a bio-metaheuristic for solving optimization problem b) Evaluation of the proposed method using multiple similarity measures and a comparative performance analysis of the proposed method.

The rest of the paper is organized as follows: In section 2 the most relevant studies are listed, in section 3 the details of the method are described, section 4 explains the experimental setup and discussion of the results, and section 5 presents the conclusion and potential future work.

## 2. RELEVANT WORK

There are many studies that focus on feature oriented camera motion estimation by tracking a template in the images. However, here we focus on the direct methods that track a planar template by optimizing the similarity between a reference and a current image. A classic example of such a direct approach toward camera motion estimation is the use of a brightness constancy assumption during motion and is linked to optical flow measurement [12]. Direct methods based on optical flow were later divided into two major pathways: Inverse

Compositional (IC) and Forward Compositional (FC) approaches [20]–[22]. The FC approaches solve the problem by estimating the iterative displacement of warp parameters and then updating the parameters by incrementing the displacement. IC approaches, on the other hand, solve the problem by updating the warp with an inverted incremental warp. These methods linearize the optimization problem by Taylor-series approximation and then solve it by least-square methods. In [23] a multi-plane estimation method along with tracking is proposed in which region-based planes are firstly detected and then the camera is tracked by minimizing the SSD (Sum of Squared Differences) between respective planar regions in 2D images. Another example of direct template tracking is [23] which improves the tracking method by replacing the Jacobian approximation in [21] with a Hyper-plane Approximation. The method in [23] is similar to our method because it embeds constraints in a non-linear optimization process (i.e. Levenberg-Marquardt [9]) although it differs from the method proposed here since the latter employs a bio-inspired metaheuristic based optimization process which maximizes the mutual information in-between images and also the proposed method does not use constraints among the planes.

## 3. METHODOLOGY

The problem that is being addressed deals with estimation of a robot's state at a given time step that satisfies the planarity constraint. Let $x(x^t, x^r) \in \mathbb{R}^6$ be the state of the robot with $x^t \in \mathbb{R}^3, x^r \in \mathbb{R}^3$ being the position and orientation of the robot in Euclidean space. Let's also consider $I, I_r$ to be the current and reference image, respectively. If the current image rotates $R \in \mathbb{SO}(3)$ and translates $t \in \mathbb{R}^3$ from the reference image in a given time step then the motion in terms of homogeneous representation $T \in \mathbb{SE}(3)$ can be given as:

$$T(x) = \begin{bmatrix} \hat{s}(x^r) & x^t \\ 0^T & 1 \end{bmatrix} \quad (1)$$

where $\hat{s}$ is the skew symmetric matrix. It is indeed this transformation that we ought to recover given the current state of the robot and reference template image.

### 3.1 Plane induced Motion

It is often the case that the robot's surrounding is composed of planar components, especially in the case of indoor navigation where most salient landmarks are likely to be planar in nature. In such cases the pixels in an image can be related to the pixels in the reference image by a projective homography H that represents the transformation between the two images [24]. If $p = [u, v, 1]^T$ be the homogeneous coordinates of the pixel in an image and $p^r = [u^r, v^r, 1]^T$ be the homogenous coordinates of the reference image then the relationship between the two set of pixels can be written as given in equation 2.

$$p \propto Hp^r \quad (2)$$

Let's now consider that the plane that is to be tracked or the plane which holds a given landmark has a normal $n_r \in \mathbb{R}^3$, which has its projection in the reference image $I_r$. In case of a calibrated camera, the intrinsic parameters, which are known, can be represented in terms of a matrix $K \in \mathbb{R}^{3 \times 3}$. If the 3D transformation between the frames is $T$, then the Euclidean homography with a non-zeros scale factor can be calculated as:

$$H(T, n_r) \propto K(R + tn_r^T)K^{-1} \tag{3}$$

### 3.2 Model-Based Image Alignment

The next step after modeling the planarity of the scene is to relate plane transformations in the 3D scene to their projected transformations in the images. For that reason a general mapping function that transforms a 2D image given a projective homography can be represented by a warping operator w and is defined as follows:

$$w(H, p^r) = \left[\frac{h_{11}u^r + h_{12}v^r + h_{13}}{h_{31}u^r + h_{32}v^r + h_{33}}, \frac{h_{21}u^r + h_{22}v^r + h_{23}}{h_{31}u^r + h_{32}v^r + h_{33}}\right]^T \tag{4}$$

If the normal of the tracked plane is known then the problem to be addressed is that of metric model based alignment or simply model based non-linear image alignment. It is the transformation $T \in \mathbb{SE}(3)$ that is to be learned by warping the reference image and measuring the similarity between the warped and the current image. Since the intensity of a pixel $I(p)$ is a non-linear function, we need a non-linear optimization procedure. More formally, the task is to learn an optimum transformation $\hat{T} = T(x)$ that maximizes the following:

$$\max_{x \in \mathbb{R}^6} \psi \left( I_r \left( w(H(\hat{T}, n_r), p_r) \right), I(p) \right) \tag{5}$$

where $\psi$ is a similarity function and $\hat{T}$ is updated as $\hat{T} \leftarrow T(x)\hat{T}$ for every new image in the sequence.

### 3.3 Similarity Measure

In order for any optimization method to work effectively and efficiently, the search space needs to be modeled in such a way that it captures the multiple optima of a function but at the same time suppresses local optima by enhancing the global optimum. It is also important that such modeling of similarity must provide enough convergence space so that the probability of missing the global optimum is minimized. This job is performed by a selection of similarity measure that is best suited for a given problem context. An often used measure is SSD (Sum of Squared Differences) that can be given as:

$$\psi_{SSD} = \sum_i^N \left( I_r(w(H, p_r)) - I(p) \right)^2 \tag{6}$$

where '$N$' is the total number of pixels in a tracked region of the image.

Similarly, another relevant similarity measure is the cross correlation coefficient of the given two data streams. Often a normalized version is used to restrict the comparison space to the range [0, 1]. The normalized cross correlation between a current image patch $I$ and a reference image patch $I_r$, with $\mu, \mu_r$ being their respective means, can be written as:

$$\psi_{NCC} = \sum_{i,j} \frac{(I_r(i,j)-\mu_r)(I(i,j)-\mu_I)}{\sqrt{(I_r^2(i,j)-\mu_r)(I^2(i,j)-\mu_I)}} \quad (7)$$

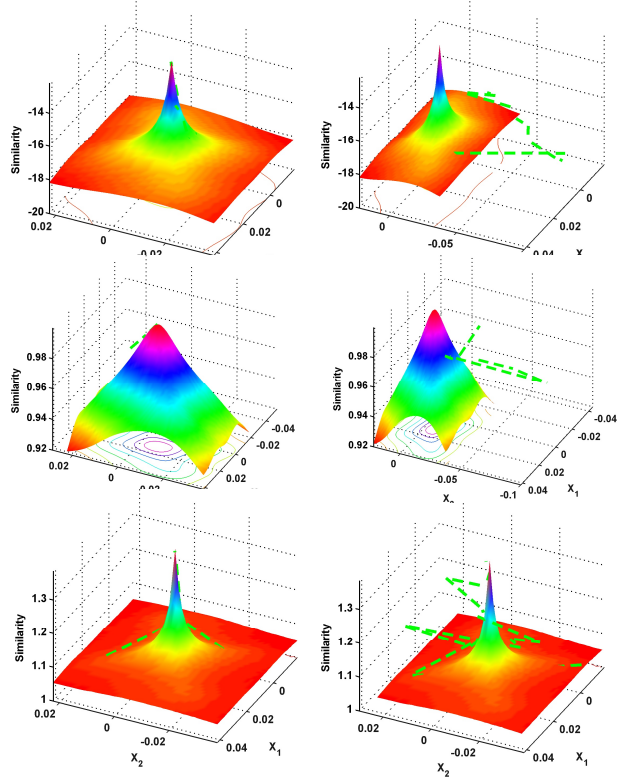

Fig. 1: Convergence surface of various similarity functions along with motion of a PSO particle on its way towards convergence depicted by green path. First Row: Sum of squared difference $(\log(\psi_{SSD})^{-1})$, Second Row: Normalized Cross Correlation $(\psi_{NCC})$, Third Row: Mutual Information $(\psi_{MI})$.

The similarity measures presented in equation 6 and 7 have the ability to represent the amount of information that is shared by the two data streams; however, as can be seen in figure 1, the convergence space and the emphasis on the global optimum need improvement. A more intuitive approach for measuring similarity among the data is Mutual Information (MI), taken from information theory, that measures the amount of data that is shared between the two input data streams [25]. The application of MI in image alignment tasks and its ability to capture the shared information have also proven to be successful [26], [27]. The reason for avoidance of MI in robotics tasks has been its relatively higher computational expense, since it involves histogram computation. However, the gains are more than the losses, so we choose to use MI as our main similarity measure. Formally, the MI between two input images can be computed as:

$$\psi_{MI} = E(I) + E(I_r) - E(I_r, I)$$

$$E(I) = -\sum_{i=0}^{N_I} \rho_I(i) \log(\rho_I(i)) \qquad (8)$$

$$E(I_r, I) = -\sum_{i=0}^{N_I} \sum_{j=0}^{N_I} \rho_\Pi(i,j) \log(\rho_\Pi(i,j))$$

where $E(I), E(I_r, I), N_I$ are the entropy, joint entropy and maximum allowable intensity value respectively.

Entropy according to Shannon [23] is the measure of variability in a random variable $I$, whereas '$i$' is the possible value of $I$ and $\rho_I(i) = \Pr(i == I(p))$ is the probability density function. The log basis only scales the unit of entropy and does not influence the optimization process. In other words, since $-\log(\rho_I(i))$ represents the measure of uncertainty of an event '$i$' and $E(I)$ is the weighted mean of the uncertainties, the latter represents the variability of random variable $I$. In the case of images we have pixel intensities as possible values of a random variable, so the probability density function of a pixel value can be estimated by a normalized histogram of the input image. The entropy for the input image is therefore a dispersion measure of the normalized histogram. This is the same in the case of joint entropy since there are two variables involved, so the joint probability density function is $\rho_\Pi(i,j) = \Pr(i == I_r(p^r) \cap j == I(p))$ where $i, j$ are possible values of image $I_r, I$ respectively. The joint entropy measures the dispersion in the joint histogram. This dispersion provides a similarity measure because when the dispersion in the joint histogram is small then the correlation among the images is strong, giving a large value of MI and suggesting that two images are aligned, while in case of large dispersions, MI would have a small value and images would be unaligned.

### 3.4 Optimization Procedure

The problem of robust retrieval of Visual Odometry (VO) in subsequent images is challenging due to the non-linear and continuous nature of the huge search space. The non-linearity is commonly tackled using linearization of the problem function; however, this approximation is not entirely general due to challenges in exact modeling of image intensity. Another route to solve the problem is to use non-linear optimization such as Newton Optimization which gives fairly good convergence due to the fact that it is based on Taylor series approximation of the similarity function. However, it requires computation of the Hessian which is computationally expensive and also it must be positive definitive for a convergence to take place.

The proposed method seeks the solution to the optimization problem presented in equation 5. In order to find absolute global extrema and not get stuck in local extrema we choose a bio-inspired metaheuristic optimization approach (i.e. PSO). Particle Swarm Optimization (PSO) is an evolutionary algorithm which is directly inspired by the grouping behavior of social animals, notably in the shape of bird flocking, fish schooling and bee swarming. The primary reason for interest in learning and modeling the science behind such activities has been the remarkable ability possessed by natural organisms to solve complex problems (e.g scattering, regrouping, maintaining course, etc.) in a seamlessly and robust fashion. The generalized encapsulation of such behaviors opens up horizons for potential applications in nearly any field. The range of problems that can be solved range from resource management tasks (e.g intelligent planning and scheduling) to real mimicked behaviors by robots. The particles in a swarm move collectively by keeping a safe distance from other members in order to avoid obstacles while moving in a consensus direction to avoid predators while maintaining a constant velocity. This results in behavior in which a flock/swarm moves towards a common goal (e.g. a Hive, food source) while intra-group motion seems random. This makes it difficult for predators to focus on one prey while it also helps swarms to maintain their course, especially in case of long journeys that are common, e.g., for migratory birds. The exact location of the goal is usually unknown as it is in the case of any optimization problem where the optimum solution is unknown. A pictorial diction of the robot's states represented as particles in an optimization process can be seen in figure 2.

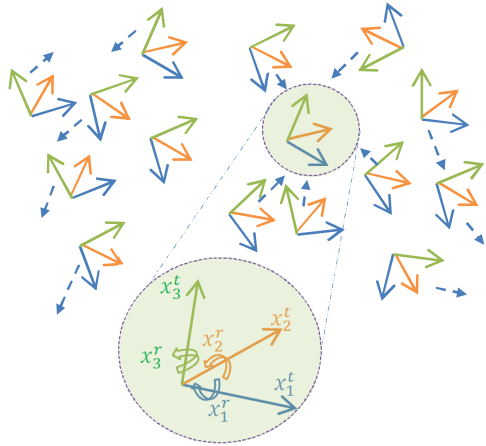

Fig. 2: A depiction of PSO particles (i.e. robot states) taking part in an optimization process. Blue arrows show the velocity of a particle and the local best solution is highlighted by an enclosing circle.

PSO is implemented in many ways with varying levels of bioinspiration reflected in terms of the neighborhood topology that is used for convergence [28]. Each particle maintains it current best position $p_{best}$ and global best $g_{best}$ position. The current best position is available to every particle in the swarm. A particle updates its position based on its velocity, which is periodically updated with a random weight. The particle that has the best position in the swarm at a given iteration attracts all other particles towards itself. The selection of attracted neighborhood as well as the force to which the particles are attracted depends on the topology being used. Generally a PSO consists of two broad functions: one for exploration and one for exploitation. The degree and extend of time that each function is performed depends again on the topology being used. A common model of PSO allows more exploration to be performed in the initial iterations while it is gradually decreased and a more localized search is performed in the later iterations of the optimization process. There can be multiple topologies, some of which are presented in figure 3. A global topology considers all the particles in the swarm and thus converges faster but potentially to a local optimum. The local best topology provides relatively more freedom and only a set of close neighbors are allowed to be attracted to the best particle in the swarm. This allows the algorithm to converge more slowly, but chances of finding the global optimum are increased. Another subtle variation to this neighborhood selection could be that it is performed dynamically, e.g. being increased over the number of iterations, making the system more explorative in the start and more exploitative in later stages of convergence so that a global optimum is achieved.

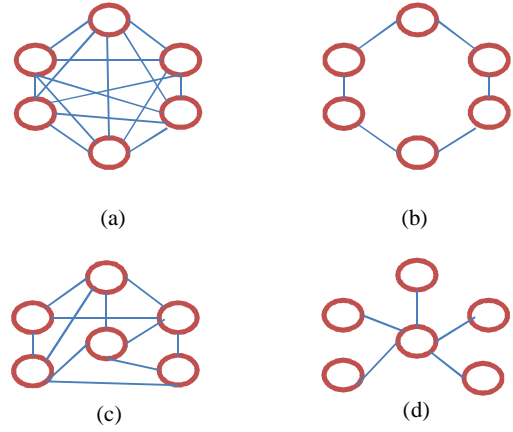

Fig. 3: Various topologies used in PSO based optimization processes. (a) Global topology uses all neighbors, (b) Circle Topology consider 2-neighbours, (c) Local Topology uses k-neighbors, and (d) Wheel Topology considers only immediate neighbors.

The process of PSO optimization starts with initialization of the particles. Each particle is initialized with a random initial velocity $v_i$ and random current position $x_i$ represented by a $k$ dimensional vector where '$k$' is the number of degrees of freedom of the solution. The search space is discretized and limited with a boundary constraint $|x_i| \leq b_i, b_i \in [b_l, b_u]$ where $b_l, b_u$ are lower and upper bounds of motion in each dimension. This discretization and application of boundary constraints helps reduce the search space assuming that the motion in between subsequent frames is not too large. After initialization, particles are moved arbitrarily in the search space to find the solution that maximizes the similarity value as given in equation 8. Each particle updates its position based on its own velocity and the position of the best particle in the neighborhood. The position and velocity update is given in equation 9:

$$x_i(t+1) = x_i(t) + v_i(t+1)$$
$$v_i(t+1) = \omega\, v_i(t) + \alpha_c \sigma_c c_i + \alpha_s \sigma_s s_i$$
(9)

where ω is the inertial weight and is used to control the momentum of the particles. When a large value of inertial weight is used, particles are influenced more by their last velocity and collisions might happen with very large values. The cognitive (or self- awareness) component of the velocity update is represented by $c_i = x_i^b - x_i(t)$ where $x_i^b$ is the personal best solution of the particle. Similarly, the social component is represented as $s_i = x_i^g - x_i(t)$ where $x_i^g$ is the best solution in the particle's neighborhood. Randomness is achieved by $\sigma_c, \sigma_s \in [0,1]$ for cognitive and social components respectively. The constant weights $\alpha_c, \alpha_s$ control the influence of each component in the update process. The final step in the PSO algorithm is evaluating against the termination criteria. There could be a range of strategies for terminating the algorithm: a) using a fixed number of iterations until improvement in the solution is observed, b) reaching a maximum number of consecutive iterations with no change in the solution, c) exceeding a threshold on the maximum similarity value. The first strategy would be fast but may fail to return a true optimum due to being deceived by a local best solution. The second strategy would be better than the first one in returning the global best solution; however it might also not guarantee the best solution. Third strategy is good in the sense that it can find the best solution, but it could make the algorithm continue for an infinite number of iterations. There is no general strategy for termination; rather, each problem situation needs to be adapted. If faster computation is the ultimate concern then the first option would be a good choice; similarly the third option would be good in situations where the best solution is to be found regardless of the time taken. A weighted combination of termination criteria proves better suited for the problem at hand.

3.5 Tracking Method

The proposed plane tracking method consists of three main steps: initialization, tracking and updating. These steps are given as follows:

1) The planar area in the image that is to be tracked is initialized in the first frame and an initial normal of the plane is provided. If the plane normal is not already known then a rough estimate of the plane in the camera coordinate frame is given. The search space of the problem is discretized and constrained within an interval. PSO is initialized with a random solution and a suitable similarity function is provided.
2) The marked region in the template image is aligned with a region in the current image and an optimum solution of the 6-dof transformation is obtained. The optimization process continues until it meets one of the following conditions: (i) max number of iterations is reached, (ii) the solution has not improved in a number of consecutive iterations, or (iii) a threshold for solution improvement is reached.
3) The global camera transformation is updated and process repeats.

## 4. EXPERIMENTAL RESULTS

In order for the system to be evaluated the experiment setup must consider the basic assumptions namely: planarity nature of the scene and small subsequent motion. The planarity nature of the scene means that there should be a dominant plane in front of the camera whose normal is either estimated by using another technique or using an approximated unit normal without scale, however the rate of convergence and efficiency is affected in the latter case. The second important assumption of the system is that the amount of motion in subsequent frames is small as large motions increase the search space significantly. In addition to this, the planar region that is to be tracked must be textured so as to provide good variance of similarity while being transformed. Keeping these assumptions in mind, the algorithm is evaluated for both simulated and real robotic transformations and results are recorded.

4.1 Synthetic sequence

The experimentation process using a benchmark tracking sequence is performed. The sequence consists of a real image with a textured plane facing the camera and its 100 transformed variations whereas the motion within the subsequent transformation is kept small. The tracking region is marked in the template image in order to select the plane and the optimization algorithm is initialized. The tracking method succeeds in capturing the motion as it can be seen in figure 4. In order to test behavior of the similarity measures, the method is repeated with all three similarity functions and error surface is analyzed which can be seen in the figure 1 which also show the path of a particle in the swarm on its way toward convergence. It was found that MI provides better convergence surface than other two participating similarity measures and hence it is used for later stages of the evaluation process.

As it could be interesting to determine whether the algorithm could cater variation in the degree of freedom, the sequence is run multiple times with different dimensions of the solution that is to be learned. The increase in the number of parameters to be learned affects the convergence rate, however the algorithm successfully converges all the variations as can be seen in the figure 5. However, with an increase in degree of freedom, the search space expands exponentially making it harder to converge in the same number of iterations as needed for lower degrees of freedom. This can be catered by multiple ways; a) increasing the overall number of iterations needed by the algorithm to converge b) increasing the number of iterations dedicated for exploration and c) putting more emphasis on the exploration by setting the appropriate inertial and social weights in equation 9. The learning of all 6 parameters of a robotic motion is a challenging task and is often reduced to 4 DOF by supplementing the two DOF from the IMU however using the optimization based approach such as presented in this paper could be used for such challenging tasks and could successfully solve the problem. It is important to note here that the proposed method only uses single plane and hence only one constraint or more formally it is an unconstrained plane segment tracking while introducing more planes and introduction of their inter-plane constraints could lead to significant increase in stability and accuracy.

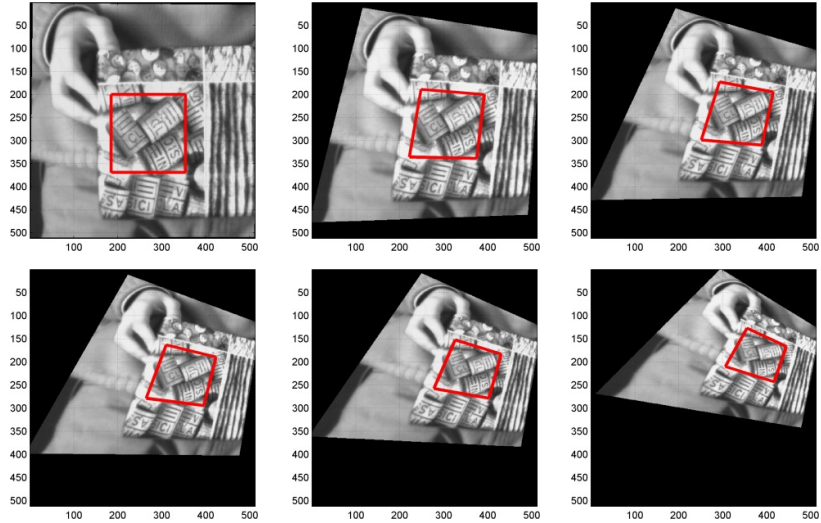

Figure 4: The result of the tracking when applied on a benchmarking image sequence with synthetic transformations.

### 4.2 Real Sequence

The proposed tracking method is also tested on a real image benchmarking sequence that is recorded by a downward looking camera mounted on a quadrocopter [30]. The sequence consists of 633 images with the resolution 752x480 recorded by flying the quadrotor in a loop. The important variable that was unavailable in case of real images is the absolute normal of the tracked plane. There could be two ways to solve this problem: using an external plane detection method to estimate the normal or using a rough estimate of the plane and leave rest to optimization process. The former approach is more preferable and could lead to better convergence rate however, to show the insensitivity of the proposed method to absolute plane normal and depth estimates, we use the latter approach for evaluation. The rest of the parameterization and initialization process is similar to simulated sequence based evaluation process as described earlier.

It could be visualized in figure 6 that even though initial transformation of the marked region was not correct and absolute normal was unknown; the tracking method learns the correct transformation over a period of time and successfully tracks the planar region.

A through error analysis is provided in the figure 7 which shows that the proposed tracking method has good tracking ability with minimal translation and rotational error when the motion is kept within the bounds of search space. A good way to keep the motion small is attained by using high frame-rate cameras.

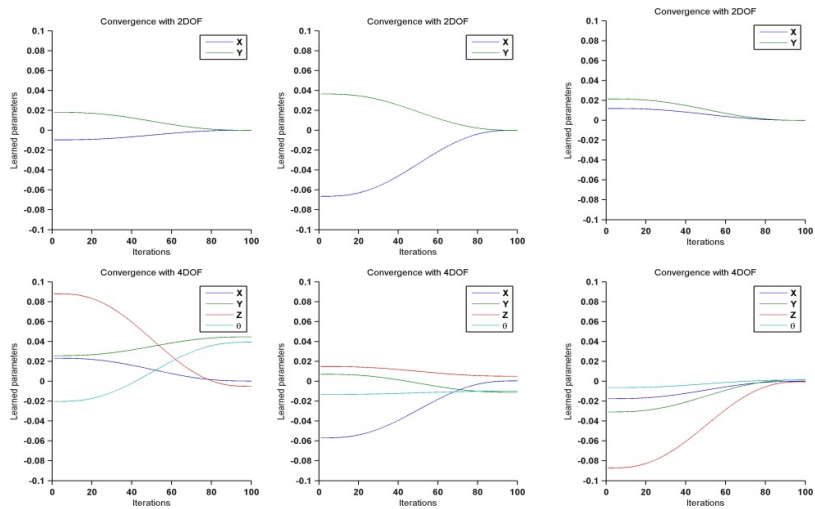

Figure 5: Convergence with variation in DOF and similarity measures. The columns represent similarity measures SSD, NCC and MI respectively and rows represent DOF 2 and 4 respectively.

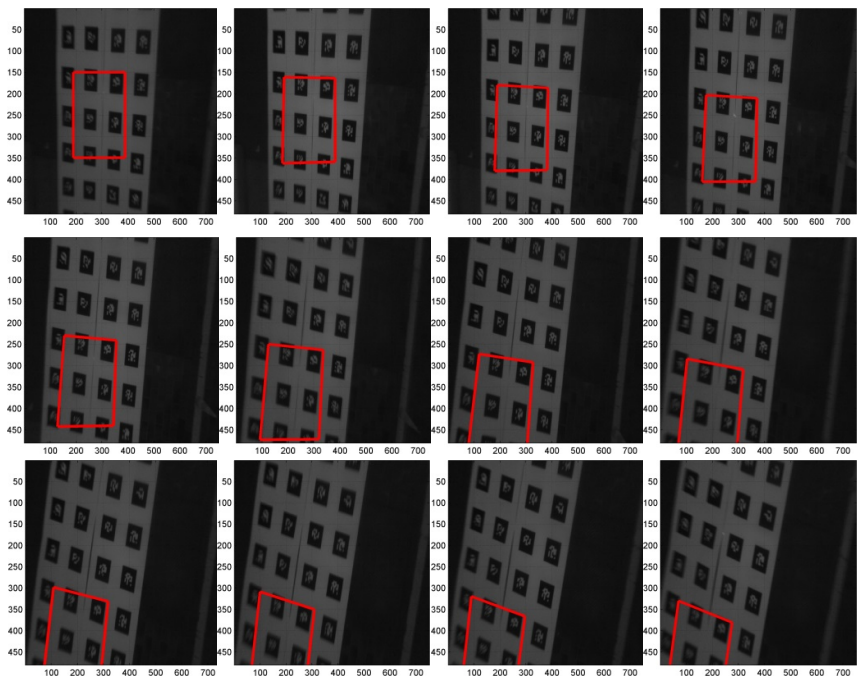

Figure 6: The result of the tracking when applied on a benchmarking image sequence with real transformations.

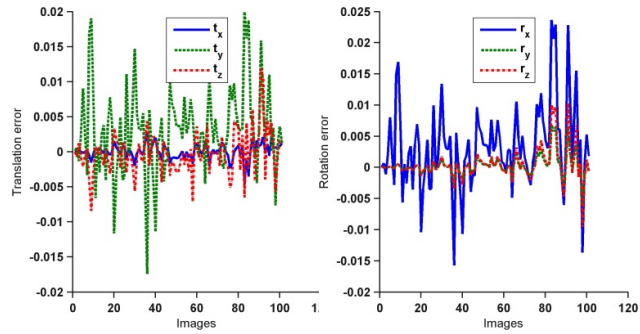

Figure 7: Transformation error for image sequence.

## 4.3 Comparative analysis

The proposed method performs an optimization based tracking so comparative analyses with variations of PSO and also with other state of the art methods help us determine its significance in real applications. In figure 8 a comparison of the multiple variations of PSO is presented. The Trelea PSO is good at converging to optimum similarity in all cases although its convergence rate is not fastest due to being explorative in nature. PSO common on the other hand finds its way quickly toward solution although it may not find global optimum due to being more exploitative in nature. A group of three state of the art plane tracking methods (IC, FC and HA) are applied on the same image sequence and a normalized root mean squared error is measured for the image sequence and the number of iterations. As it can be visualized from the figure 9 that the algorithm successfully beats IC and HA while it nears the performance of the forward compositional approach.

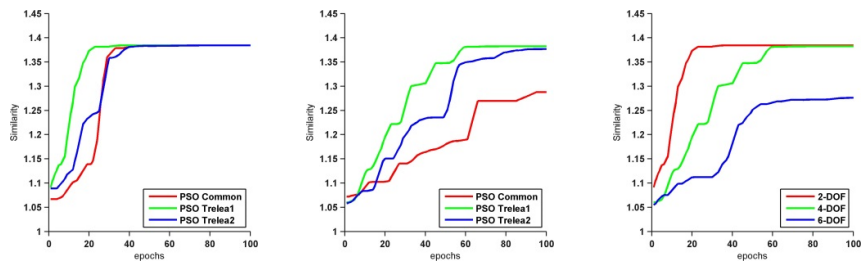

Figure 8: comparison of the convergence rate of variations of PSO along with convergence behavior of best performing version of PSO with different degree of freedom.

It can be noted that the IC and HA misses the track of the plane after $40^{th}$ iteration, most probably due to intensity variation that is introduced in the sequence for which Taylor series approximation failed to capture the intensity function. As a comparison if we check the performance of the methods with different degree of freedom (see figure 10), we can see that the proposed PSO-Track method performs decently.

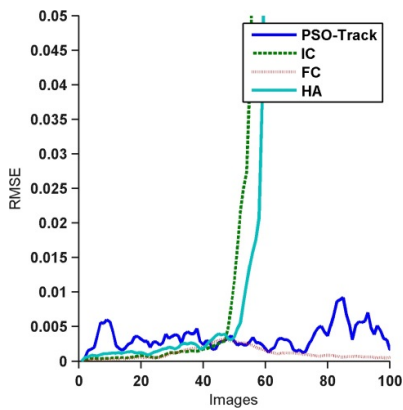 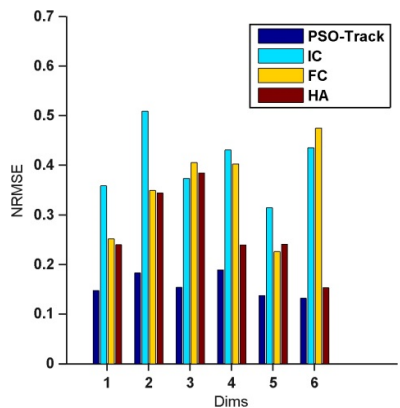

Figure 9: Performance comparison of various tracking methods.

Figure 10: Performance comparison of various tracking methods with variation in DOF.

## 5. CONCLUSIONS

In this paper, we presented a novel approach toward solving camera tracking by tracking a projection of the plane. A non-linear image alignment is adopted and correct parameters of the transformation are recovered by optimizing the similarity between the planar regions. A through comparative analysis of the method over simulated and real sequence of images reveal that the proposed method has ability to track planar surfaces in when the motion within the frames is kept small. The insensitivity of the method toward intensity variations as well as to unavailability of true plane normal is also tested and algorithm has been found resilient to such environmental changes.